%% file: root_arxiv.tex
\documentclass[letterpaper, 10 pt, conference]{ieeeconf}

\IEEEoverridecommandlockouts

\usepackage[table]{xcolor}
\usepackage{graphicx}
\usepackage{booktabs}
\usepackage{multirow}
\usepackage{newtxtext}
\usepackage{newtxmath}
\usepackage{amsmath} 
\usepackage{balance}
\usepackage{cite}
\usepackage{xspace}
\usepackage{tcolorbox}
\usepackage{textcomp}
\usepackage{subcaption}
\usepackage{pifont}

\newcommand{\eg}{\emph{e.g.}}
\newcommand{\ie}{\emph{i.e.}}

\newcommand{\dataset}{\texttt{GroundedPlanBench}\xspace}
\newcommand{\method}{\texttt{V2GP}\xspace}
\newcommand{\taskname}{\texttt{grounded planning}\xspace}
\definecolor{fftgray}{RGB}{235,235,235}
\title{\LARGE \bf
Spatially Grounded Long-Horizon Task Planning in the Wild
}

\author{Sehun Jung$^{1\dagger}$, HyunJee Song$^{1\dagger}$, Dong-Hee Kim$^{1}$, Reuben Tan$^{2}$, Jianfeng Gao$^{2}$, Yong Jae Lee$^{3}$, and Donghyun Kim\textsuperscript{1\ding{41}} 
\thanks{$\dagger$ indicates first authors with equal contributions.}
\thanks{$^{1}$Korea University, $^{2}$Microsoft Research $^{3}$University of Wisconsin-Madison}
\thanks{* The work was supported the Institute of Information \&communications Technology Planning \& Evaluation (IITP) grant funded by the Korea government(MSIT) (No. RS-2025-25439490). (\ding{41} Corresponding author: \texttt{d\_kim@korea.ac.kr}). }
}

\begin{document}
\maketitle
\thispagestyle{empty}
\pagestyle{empty}


\begin{abstract}
Recent advances in robot manipulation increasingly leverage Vision–Language Models (VLMs) for high-level reasoning, such as decomposing task instructions into sequential action plans expressed in natural language that guide downstream low-level motor execution. 
However, current benchmarks do not assess whether these plans are spatially executable, particularly in specifying the exact spatial locations where the robot should interact to execute the plan, limiting evaluation of real-world manipulation capability.
To bridge this gap, we define a novel task of \taskname and introduce \dataset, a newly curated benchmark for \textbf{spatially grounded long-horizon action planning in the wild}. \dataset jointly evaluates hierarchical sub-action planning and spatial action grounding (\emph{where to act}), enabling systematic assessment of whether generated sub-actions are spatially executable for robot manipulation.
We further introduce Video-to-Spatially Grounded Planning (\method), an automated data generation framework that leverages real-world robot video demonstrations to improve spatially grounded long-horizon planning. Our evaluations reveal that spatially grounded long-horizon planning remains a major bottleneck for current VLMs. 
Our results demonstrate that \method\ provides a promising approach for improving both action planning and spatial grounding performance, validated on our benchmark as well as through real-world robot manipulation experiments, advancing progress toward spatially actionable planning.
\end{abstract}

\section{Introduction}

\begin{figure}[t]
    \centering
    \begin{subfigure}{\columnwidth}
        \centering
        \includegraphics[width=\columnwidth]{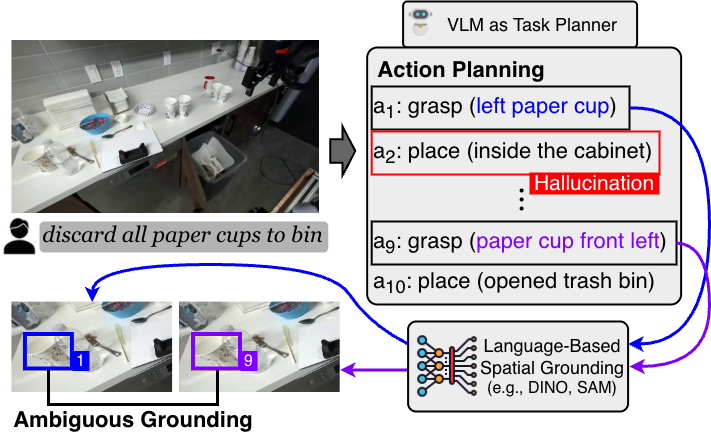}
        \vspace{-6mm} 
        \caption{Limitation of prior decoupled task planning \& spatial grounding.}
        \label{fig:teaser_a}
    \end{subfigure}
    
    \vspace{0.3cm} 
    
    \begin{subfigure}{\columnwidth}
        \centering
        \includegraphics[width=\columnwidth]{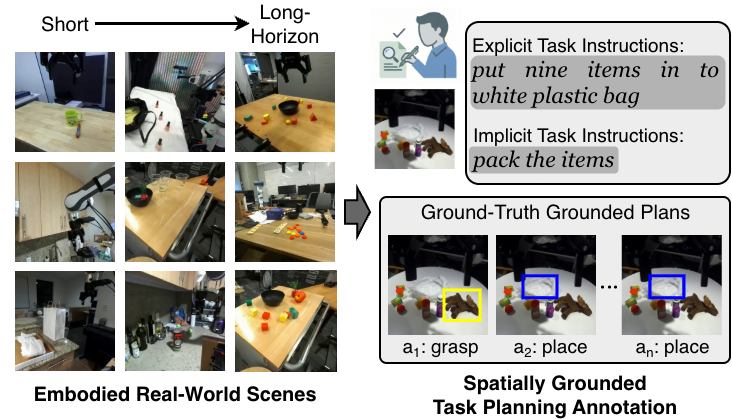}
        \vspace{-6mm} 
        \caption{Construction of \dataset.}
        \label{fig:teaser_b}
    \end{subfigure}
    \vspace{-4mm} 
    \caption{Motivation of \taskname.
    (a) VLM-as-Planner decomposes high-level instructions into natural language sub-actions, which are then grounded by separate perception modules. However, the lack of explicit spatial specification often leads to ambiguous action grounding.
    (b) Our \dataset jointly annotates and evaluates hierarchical sub-action planning and spatial grounding under both explicit and implicit instructions.
    }
    \vspace{-3mm} 
    \label{fig:teaser}
\end{figure}

Recent advances in Vision–Language Models (VLMs), trained on large-scale web data, have driven a paradigm shift in robotics toward general-purpose agents capable of robust perception, reasoning, and generalization for interaction and manipulation in open-world environments. A dominant paradigm uses VLMs to directly map visual observations and language instructions to low-level robot motor signals, known as Vision–Language–Action (VLA) models~\cite{openvla,octo,magma}. While promising, these end-to-end VLAs often struggle with generalization across diverse scenes and tasks~\cite{fsd,embodeidr1}. Consequently, VLM-as-Perception approaches~\cite{embodeidr1,fsd} decouple high-level reasoning from low-level motor control, using VLMs for spatial perception to guide downstream execution (\eg, motion planning), thereby improving generalization in robotic manipulation. However, such methods typically lack the hierarchical reasoning capabilities necessary for long-horizon task decomposition and planning.
In parallel, the VLM-as-Planner~\cite{codeaspolicy,cotvla,moka} exploits the reasoning ability of VLMs to decompose complex instructions into sequences of low-level actions expressed in natural language. 

In this paper, we examine a key limitation of VLM-as-Planner: the generated sub-actions, expressed in natural language, are often ambiguous or hallucinated, resulting in plans that are not physically or spatially executable. 
However, systematically evaluating these issues remains challenging due to the lack of comprehensive benchmarks. Existing evaluation benchmarks generally fall into two categories, each with notable limitations: (1) VLM for task planning~\cite{robovqa,embodiedbench,vlabench,robobrain} benchmarks evaluate the ability to decompose high-level instructions into low-level action plans.  While effective for abstract planning, these benchmarks are often simulator-based, limiting exposure to diverse real-world scenes and tasks, and frequently lack explicit evaluation of spatial grounding during planning.  Consequently, they measure what actions to perform but not where to execute them, making it challenging to verify spatial feasibility. As illustrated in Fig.~\ref{fig:teaser_a}, the natural language sub-actions generated by the VLM-as-Planner paradigm often fail to accurately specify where to act due to inherent ambiguities or hallucination in linguistic descriptions,

(2) Visual Question Answering (VQA) embodied benchmarks~\cite{robospatial,embodeidr1} have been proposed to evaluate spatial reasoning capability (\eg, grounding an object given a text query), which does not natively evaluate a mechanism for decomposing long-horizon tasks. 

We define a novel task of \taskname, where the goal is to perform \textbf{spatially grounded and long-horizon action planning} for task completion in the wild. To bridge this gap, we introduce \dataset, a new benchmark designed to
rigorously assess two critical dimensions: (1)  hierarchical action planning, which requires deriving appropriate actions across diverse scenes and tasks, spanning short- to long-horizon objectives based on both explicit and implicit user instructions, and (2) spatial action grounding, which requires the model to precisely identify where to interact with objects for each action step. Unlike prior benchmarks, by jointly evaluating action planning and spatial grounding, our benchmark assesses whether VLM-generated plans correspond to physically feasible robot manipulation behaviors (Fig.~\ref{fig:teaser_b}). 
\dataset is built from 308 diverse real-world embodied scenes drawn from DROID\cite{droid}, each annotated with manually curated action plans and spatial grounding under both implicit and explicit instructions, yielding 1K evaluation episodes.

Finally, to address the challenges introduced by this benchmark, we propose Video-to-Spatially Grounded Planning (\method), an automated training data generation framework that improves spatially grounded planning of VLM-as-Planner by learning from real-world robot video demonstrations. Our approach leverages the video understanding and grounding capabilities of VLMs, connecting them to our task to extract structured sub-action plans from long-horizon instructions while spatially grounding robot–object interactions. This enables the generation of plans that are both logically consistent and physically executable.

Using our benchmark, we evaluate closed-source and open-source VLMs on \taskname. We find that (1) planning under implicit or long-horizon instructions in diverse real-world scenes remains challenging, (2) spatial grounding of actions (``where to act'') in current VLMs is limited even with strong grounding models (\eg, SAM3~\cite{sam3}), and (3) \method, substantially improves planning and grounding capabilities, as confirmed by our benchmark and real-world robot manipulation experiments.

Our main contributions are summarized as follows:
\begin{itemize}
\item We introduce a benchmark \dataset that jointly evaluates hierarchical action planning and spatial action grounding in the wild, enabling evaluation of whether VLM-generated plans are spatially feasible for robot manipulation.

\item We propose an automated training data collection framework from robot video demonstrations that leverages VLM video understanding and grounding capabilities to automatically discover hierarchical sub-action plans and spatially grounded robot–object interactions from long-horizon demonstrations, improving spatially grounded planning in VLM-as-Planner.
\item Through systematic evaluation of closed-source and open-source VLMs, we show that spatially grounded long-horizon planning remains a major challenge in current VLM-as-Planner, and demonstrate that \method significantly improves planning and grounding performance, validated both on our benchmark and through real-world robot manipulation experiments.
\end{itemize}

\section{Related Works}

\noindent\textbf{VLMs for Robot Manipulation.}
Recent work utilizes Vision–Language Models (VLMs)~\cite{qwen3vl,internvl35} in robotic manipulation~\cite{openvla,octo,magma,fsd,embodeidr1} to enhance generalization capabilities. By integrating pre-trained VLMs into action decision-making pipelines, robots can leverage advanced spatial reasoning and planning capabilities learned from large-scale multimodal pre-training.
End-to-end approaches using VLM directly output low-level robot actions (\eg, $\Delta x, \Delta \theta$) from visual input and task instructions. However, these models often fail to consistently translate rich perceptual understanding into effective robotic actions across diverse scenes and tasks, thereby limiting their generalization capabilities, as demonstrated in~\cite{fsd}. To bridge the gap between perception and action, VLM-as-Perception approaches \cite{moka,fsd,embodeidr1} decouple high-level spatial perception and reasoning from low-level motor execution. These frameworks leverage VLMs for semantic spatial perception and generate intermediate representations (e.g., keypoints), which are then passed to downstream controllers such as motion planners for execution. However, this process is typically evaluated on single-step reasoning scenarios and is not designed to handle long-horizon tasks that require structured task planning. Another research direction leverages VLM-as-Planner~\cite{codeaspolicy, cotvla, moka}, utilizing their reasoning capabilities to break down long-horizon instructions into a sequence of low-level actions in language descriptions. These sub-actions serve as prompts for subsequent VLM-based perception or manipulation modules for action execution. A limitation of existing task planners, however, is their reliance on text-based descriptions, which often lack the spatial grounding necessary for accurate object interaction or result in hallucinated action plans that are not physically executable, as shown in Fig.~\ref{fig:teaser_a}.
 
\noindent\textbf{Evaluating VLMs in Embodied Tasks.}
Existing frameworks for evaluating robotic reasoning in VLMs can be broadly categorized into two groups. 
Some benchmarks (\eg, EmbodiedBench \cite{embodiedbench}, VLABench \cite{vlabench}) utilize simulators to evaluate long-horizon planning. Other benchmarks involve embodied spatial grounding benchmarks, particularly those centered on Embodied VQA \cite{robospatial}. Although these benchmarks are effective for object grounding, they are often insufficient for assessing the multi-step reasoning and planning logic necessitated by long-horizon tasks. To evaluate task-planning performance, RoboVQA \cite{robovqa} and RoboBrain \cite{robobrain} adopt VQA-style protocols that verify whether a model correctly selects the next sub-task in a sequence. However, taken together, existing evaluation protocols reveal several gaps in comprehensive evaluation: (1) language-based task planning is often ambiguous, which can lead to failures in robot manipulation, particularly in long-horizon scenarios; (2) simulation-based scenarios often fail to capture the diversity of real-world environments; and (3) they frequently lack evaluations for spatial grounding within sub-tasks, leaving the ``where to interact'' unaddressed.
In this paper, we introduce \dataset, a benchmark designed to evaluate the intersection of high-level task planning and spatial grounding, not only what actions to perform but also where interactions should occur. By covering a wide range of task horizons and diverse environments, \dataset provides an evaluation protocol that closely correlates with robot manipulation in the real world.

\begin{figure}[t]
    \centering
    \begin{subfigure}{\columnwidth}
        \centering
        \includegraphics[width=0.95\columnwidth]{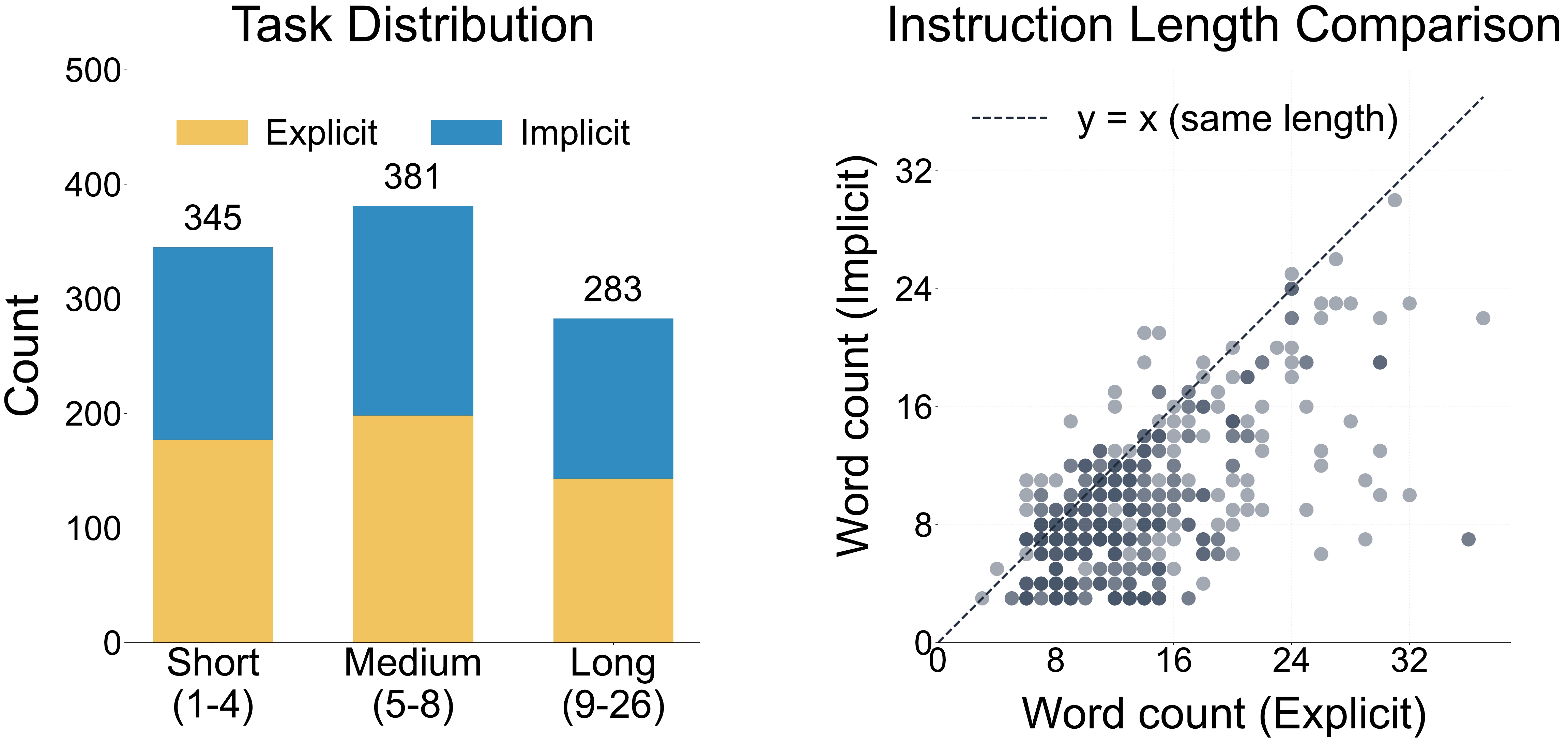}
        \vspace{-2mm} 
        \caption{ (Left) Dataset statistics. (Right) Word counts between explicit and implicit instructions. Each point represents the paired implicit and explicit instruction length for a given video.}
        \label{fig:test_statistics}
    \end{subfigure}
    
    \vspace{2mm}
    
    \begin{subfigure}{\columnwidth}
        \centering
        \includegraphics[width=\columnwidth]{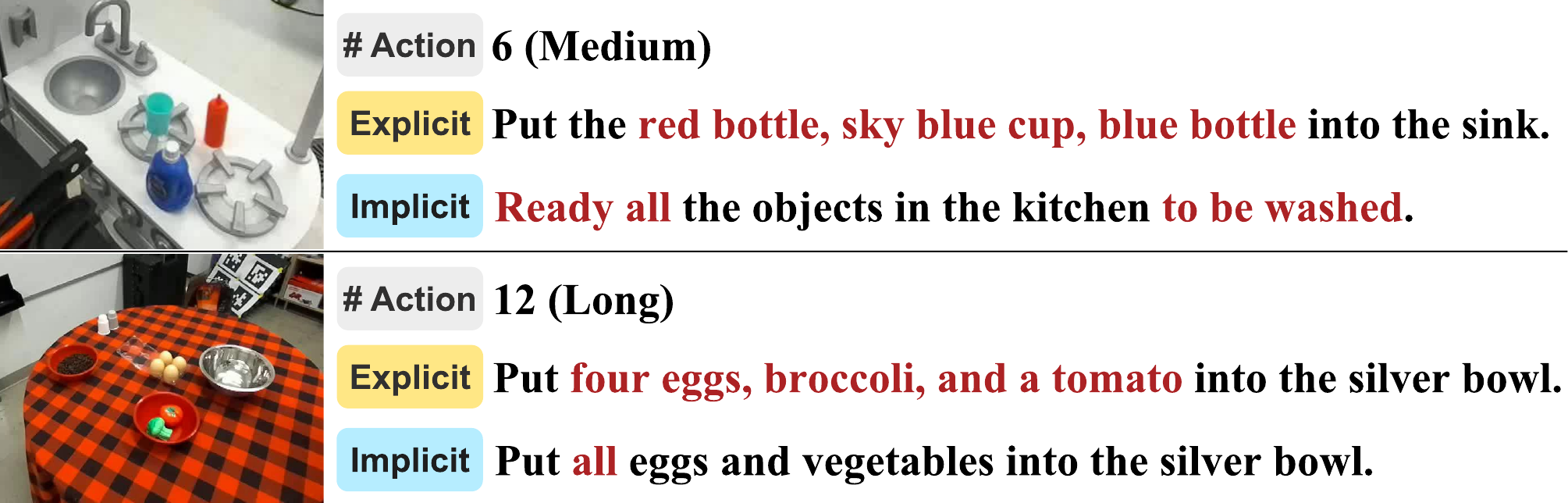}
        \vspace{-4mm} 
        \caption{Examples of paired explicit and implicit instructions.}
        \label{fig:inst_example}
    \end{subfigure}
    \vspace{-4mm} 
    \caption{ Task distribution and instruction types in \dataset.
    }
    \vspace{-4mm} 
    \label{fig:stat_dataset_example}
\end{figure}

\noindent\textbf{Learning Robot Manipulation from Video Demonstrations.}
Robot video demonstrations offer a rich data source for capturing the granular details of each sub-action required to accomplish a high-level goal. Using these videos, RoboVQA \cite{robovqa} employs human annotators to derive VQA question-answer pairs specifically designed for task planning evaluation. NILS~\cite{nils} leverages VLMs to automatically generate language instructions from videos, which are subsequently used to train language-conditioned policy networks. In contrast, we employ VLMs to extract structured sub-action plans from real-world videos~\cite{droid}, together with the associated spatial grounding that identifies where the robot should interact, linking these elements to the given high-level task instructions. The resulting annotations are then used as training data to improve VLMs for spatially grounded long-horizon task planning.

\section{Benchmark Overview}
\label{sec:benchmark}
We introduce \dataset, a benchmark designed to evaluate the spatially grounded action planning capabilities of VLM-as-Planner in unconstrained environments. This benchmark encompasses varied task horizons and both implicit and explicit instructions across a wide array of scenes and objects, as illustrated in Fig.~\ref{fig:stat_dataset_example}. 

\noindent\textbf{Task Definition: \texttt{Grounded Planning}.} %
Given an image $I$ and corresponding high-level instruction $H$, the grounded plan $a^{(I, H)}$ for executing the task can be decomposed into a sequence of $N$ sub-steps, $a^{(I, H)} = \left[a_1, a_2, \ldots, a_N \right]$, where each $a_i$ denotes a primitive action at the $i$-th substep. Motivated by~\cite{seedo}, we treat the `grasp' and `place' as the essential atomic primitives for general-purpose robotic manipulation, and additionally define `open' and `close' as separate primitives for functional interactions with articulated objects (\eg, drawers and cabinets). In contrast to prior task planning literature, we explicitly ground each sub-action ($a_i$) in the plan with spatial signals (\ie, bounding boxes and points), thereby removing the localization ambiguity inherent in natural language instructions. To this end, we prompt the VLM task planner to generate spatially grounded action plans using prompts of the following form (simplified here and further optimized for each VLM) in Fig.~\ref{fig:judge_instruction}:

\begin{figure}[h]
\centering
    \vspace{-3mm}
    \begin{tcolorbox}[
        colback=gray!5,
        colframe=black,
        boxrule=0.5pt,
        arc=2mm,
        left=2pt, right=2pt, top=2pt, bottom=2pt,
        ]
        \scriptsize\ttfamily
        \setlength{\tabcolsep}{4pt}
        \renewcommand{\arraystretch}{1.05}
        
        \begin{tabular}{p{0.95\linewidth}}
        Input: \\
        - One observation image \{$I$\} \\
        - One high-level instruction \{$H$\} \\
        Goal: Generate a spatially grounded action plan using ONLY: 
        \{open, grasp, place, close\}. \\
        Each action must operate on exactly one grounded entity. \\
        Primitive spec: \\
        - open(target\_text, bbox) \\
        - close(target\_text, bbox) \\
        - grasp(target\_text, bbox) \\
        - place(target\_text, point) \\
        \end{tabular}
    \end{tcolorbox}
    \vspace{-3mm}
    \caption{An example of a simplified instruction prompt for spatially grounded planning.}
    \vspace{-2mm}
    \label{fig:judge_instruction}
\end{figure}

\noindent\textbf{Data Selection and Annotation.}
Fig.~\ref{fig:teaser_b} illustrates an overview of the overall pipeline. To evaluate the ability of VLMs to perform spatially grounded action planning in real-world settings, we build upon the DROID dataset~\cite{droid}, a large-scale embodied robotics resource containing demonstrations with substantial object diversity across 564 scenes and 86 tasks. To construct our test benchmark, we sample $308$ videos spanning short-, medium-, and long-horizon demonstrations. From the first frame of each video, human annotators define a scene-executable task and provide both explicit (concrete and detailed) and implicit (abstract) instructions, along with the corresponding spatially grounded action plan $\left[a_1, a_2, \ldots, a_n \right]$. Each grasp, open, and close action is grounded using a bounding box over the target object, while the place action is grounded with a bounding box indicating the intended destination or receptacle region. We then group episodes by annotated action horizon: plans with 1–4 actions are categorized as Short, 5–8 as Medium, and 9–26 as Long, totaling 1,009 evaluation episodes. Fig.~\ref{fig:stat_dataset_example} represents the overall statistics of \dataset.

\begin{figure*}[t]
  \centering
  \includegraphics[width=\textwidth]{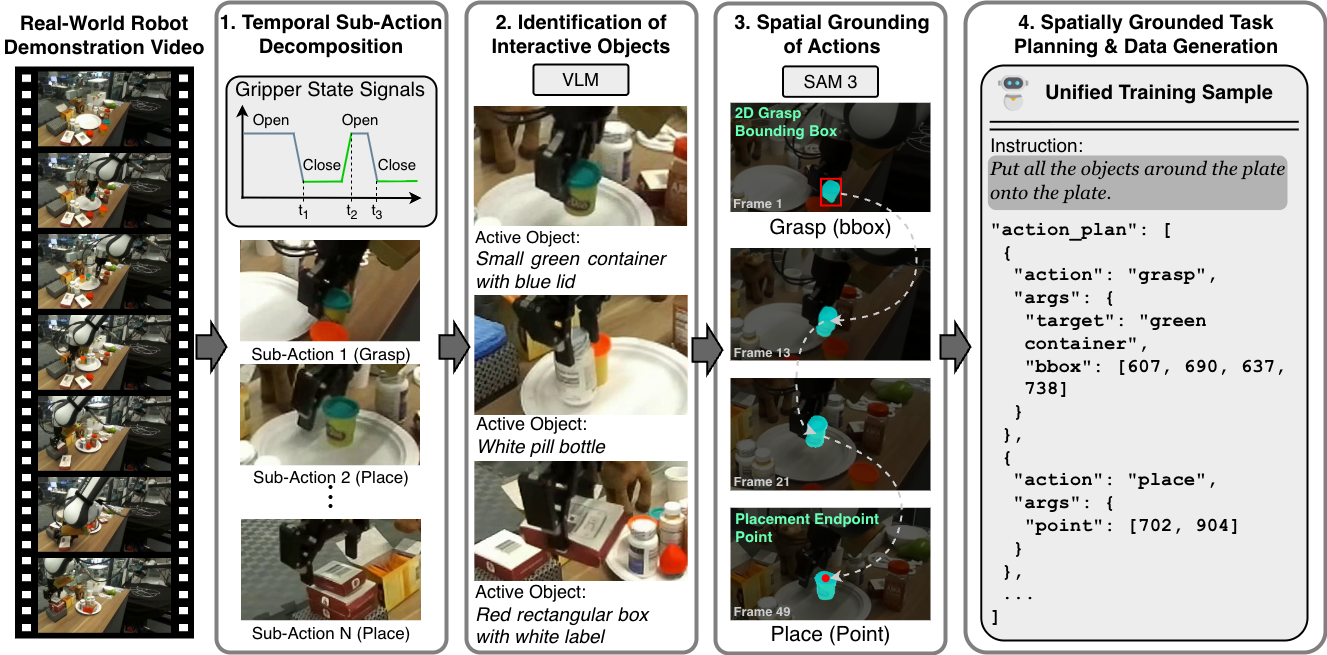}
  \caption{\method is a training data generation framework designed to enhance spatially grounded sub-action plans from real-world robot demonstration videos through the stages: (1) Temporal sub-action decomposition, where gripper state signals segment demonstrations into sub-action units; (2) Interactive object identification, where a VLM analyzes each segment to identify the actively manipulated objects; (3) Spatial grounding of actions, where SAM3 localizes target objects and placement endpoints using bounding boxes and points; and (4) Spatially grounded task planning, which integrates the grounded sub-action primitives with explicit and implicit task instructions and spatial grounding. The collected  data are used to fine-tune VLM-as-Planners, enhancing both hierarchical task planning (what to do) and spatial grounding (where to act) for long-horizon tasks.}
  \label{fig:method}
\end{figure*}

\noindent\textbf{Evaluation Metric.}
To rigorously evaluate the grounded planning capabilities of VLMs on \dataset, we adopt metrics that jointly measure sequential planning accuracy and spatial precision. We assess each action step, requiring both the correct ordering of primitives and precise spatial grounding. 

Since a grasp primitive is always followed by a place primitive in ground-truth plan, we evaluate grasp–place as a single atomic pair. Specifically, we verify whether the Intersection over Union (IoU) between predicted and ground-truth (GT) grasp bounding boxes exceeds a predefined IoU threshold ($\tau_g=0.5$), and verify whether the predicted placement point ($\hat{p}_i $) falls within the corresponding GT placement bounding box ($B_i^{p}$). Similarly, for open and close primitives, we verify whether the IoU between the predicted and GT bounding boxes exceeds a predefined threshold ($\tau_d=0.5$). In summary, we define the sub-action success as:
\vspace{-1mm}
\[
\mathrm{Succ}_i =
\begin{cases}
    \mathbf{1}\!\left[
    \mathrm{IoU}_i \ge \tau_g \;\wedge\;
    \hat{p}_i \in B_i^{p}
    \right],
    & \text{grasp--place}, \\[6pt]
    \mathbf{1}\!\left[
    \mathrm{IoU}_i \ge \tau_d
    \right],
    & \text{open/close}.
\end{cases}
\]

Additionally, while task planning often requires a specific sequence of actions (\eg, ``move cup first and then move a spoon''), certain tasks allow for multiple valid execution sequences, such as ``move all objects into the pot.'' To account for this flexibility, we define an Unordered Action Group, in which the relative ordering of actions does not affect overall task validity.
Within each unordered block, any permutation of the GT actions is considered correct.
Based on these criteria, we define the following evaluation metrics:
\begin{itemize}
    \item 	Action Recall Rate (ARR): ARR measures the proportion of generated actions that match the GT sub-actions, without considering their sequential orders.
    
    \item 	Task Success Rate (TSR): A task is considered successful if and only if all actions in the sequence are correctly planned and spatially grounded.
\end{itemize}

\section{Method}
To address the task of \taskname, we propose a new post-training approach with Video-to-Spatially Grounded Planning (\method), an automated data-generation framework designed to enhance the spatially grounded long-horizon planning capabilities: (1) \method extracts spatially executable plans aligned with high-level task instructions from the real-world robot demonstration videos in the wild, (2) Using the extracted data, we then finetune a pre-trained VLM to perform spatially grounded planning. Specifically, we leverage DROID~\cite{droid}, which contains human teleoperation annotated with high-level task instructions, robot end-effector trajectories, and gripper actions. By analyzing these videos (using data splits disjoint from \dataset), V2GP decomposes demonstrations into sub-action sequences and identifies interaction points (e.g., grasp-place), producing sequences of spatially grounded executable sub-action plans.
\method consists of the following four stages, as shwon in Fig.~\ref{fig:method}:

\noindent\textbf{Temporal Sub-Action Decomposition.}
The first stage decomposes an original video into a temporal sequence of sub-actions. We determine the temporal boundary of each sub-action using the recorded gripper state signal. Specifically, we detect the frame where the gripper begins to close and the subsequent frame where it is fully open again, and define the interval between them as a single manipulation segment, since a close–open cycle implies that the robot interacted with an object. By identifying these transitions, we convert a continuous video $v$ into a sequence of sub-action segments,	$v = [(s_1, e_1), (s_2, e_2), \ldots, (s_n, e_n)]$, where $(s_i, e_i)$ denotes the start and end frames of a $i$-th gripper close–open operation, representing steps required to accomplish the high-level goals.

\noindent\textbf{Identification of Interactive Objects.}
In the original DROID task instructions, interactive objects are often described ambiguously or at a high level (\eg, ``an object''). To resolve this, we leverage visual evidence from the video to identify which object the robot actually manipulates. Specifically, for each decomposed sub-action segment ($s_i, e_i$) extracted from the original video, we prompt a VLM (e.g., Gemini~\cite{gemini}, Qwen-VL~\cite{qwen3vl}) to infer a detailed, visually grounded description of the manipulated object (e.g., ``blue rectangular box on the counter'') along with its semantic category (e.g., ``box''). This process yields more precise object identifiers and improves the reliability of spatial grounding in later stages by establishing the ``what'' before determining the ``where.''

\noindent\textbf{Spatial Grounding of Actions.}
We aim to associate the identified object with precise 2D spatial locations that specify where the action should be executed. To this end, we track the manipulated object using SAM3 [18], prompted with textual object cues from two sources: (1) the object expression in the original instruction and (2) the visually grounded identifier and the object category from the previous stage. These complementary cues guide SAM3 to localize and track the active object across the temporal span of each sub-action, mitigating detection sensitivity to variations in prompts. This process produces spatial representations required for physical feasibility, including a 2D bounding box for the target object (\ie, $grasp(bbox)$, $open(bbox)$, and $close(bbox)$) and a placement endpoint (\ie, $place(point)$) indicating the intended destination. By grounding the spatial location of each sub-action $a_i$ within the visual scene, the resulting plans provide explicit where-to-interact signals, enabling direct translation from high-level reasoning to low-level motor execution.

\noindent\textbf{Spatially Grounded Task Planning.}
The final stage integrates the identified semantic objects and their corresponding spatial coordinates within each temporal segment into a unified training sample. We aggregate the sequence of $n$ detected sub-actions in the original video into a structured spatially grounded plan $a = [a_1, a_2, \ldots, a_n]$, where each atomic unit $a_i$ is represented as a visually grounded tuple conditioned on its primitive type. Specifically, in grasp-place interactions, $a_i = grasp(bbox)$ is followed by corresponding $a_{i+1} = place(point)$. For articulated interactions, $a_i = open(bbox)$ or $a_i = close(bbox)$. This resulting action sequence is paired with the original DROID instructions. To enhance the model’s ability to generate executable plans from implicit instructions, we further generate implicit variants of the explicit DROID instructions using Gemini to rewrite them into more abstract task formulations (\eg, ``tidy up the table”). Consequently, the training data includes both explicit and implicit instructions, enabling the model to generate scene-grounded plans even when instructions are implicit, as in real-world interactions.

\noindent\textbf{Filtering and Refinement.}	
Despite the automation of the \method framework, the generated samples may contain noise due to demonstrator mistakes (\eg, failed grasps, dropping objects) as well as detection failures from SAM3 or the VLM. To mitigate such noise, we verify that the tracking results maintain consistent object identities throughout the entire video. 
Furthermore, when the generated plan omits some actions originally specified in the instruction due to detection or tracking failures, we revise the instruction to describe only the successfully sub-actions, thereby preserving consistency between instruction and the corresponding plan. 

Finally, our training dataset comprises 34,646 samples with action sequence lengths of 1–4, 4,368 with 5–8, and 4,448 with 9–26, totaling 43K samples. With these training data, we fine-tune the Qwen3-VL~\cite{qwen3vl} model with LoRA~\cite{lora}.

\section{Experiments}
	
\subsection{Experimental Settings}
\noindent\textbf{Baselines.} We evaluate baselines across the following two paradigms:
(i) \textit{End-to-End Spatially Grounded Planning}: In this configuration, VLMs directly generate action sequences alongside their corresponding spatial grounding (i.e., specific bounding boxes and points). We evaluate several state-of-the-art models in this category, including Gemini-3-Flash, Gemini-2.5-Flash\cite{gemini25}, Qwen3-VL~\cite{qwen3vl}, and InternVL3.5~\cite{internvl35}.
\noindent (ii)\textit{ Decoupled Task Planning and Spatial Grounding}: This scenario evaluates VLMs as high-level task planners that output natural language plans. These plans are then spatially grounded by SAM3~\cite{sam3} and the specialized VLM, Embodied-R1~\cite{embodeidr1}, where SAM3 provides segmentation-based bounding boxes and Embodied-R1 predicts interaction points. This two-stage design leverages complementary strengths: general-purpose VLMs for task planning and grounding models for spatial localization. We investigate whether this decoupled approach leads to improved task planning that remains spatially grounded, compared to direct grounded planning. To assess modular performance, we employ GPT 5.2 and Qwen3-VL as high-level task planners. We evaluate three distinct grounding configurations: (1) employing Set-of-Mark (SoM)~\cite{yang2023set} to provide visual context to the planner, (2) using SAM3 for grasp and Embodied-R1 for placement (denoted as ``SAM3 + ER1''), and (3) using Embodied-R1 for both grasp and placement to translate the planners’ natural language plans into spatially grounded coordinates (denoted as ``ER1'').

\input{tables/results3.tex}

\begin{figure*}[t]
  \centering
  \includegraphics[width=0.85\linewidth]{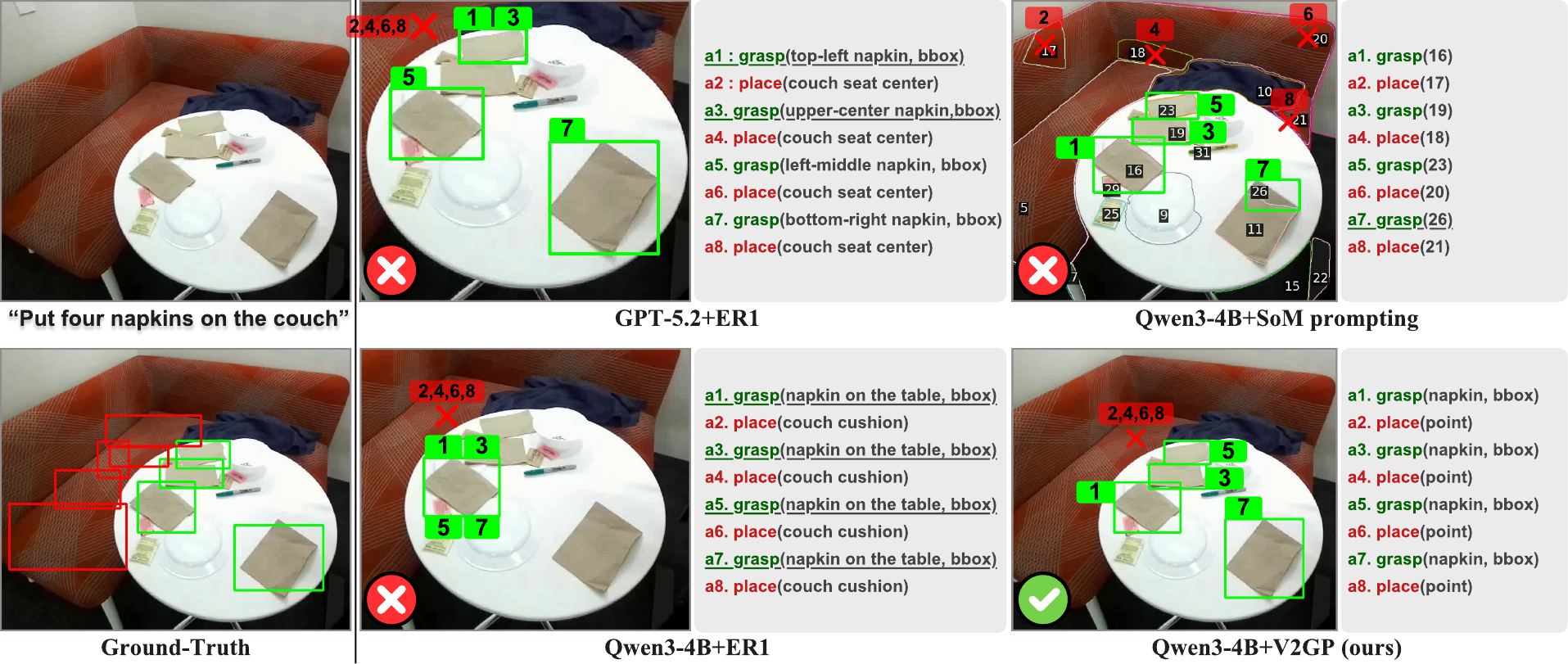}
  \vspace{-2mm}
  \caption{Visualization of decoupled task planning and spatial grounding, where \underline{underlined actions} are incorrectly grounded due to semantically similar objects (\eg, identical napkins). In contrast, \method finds the correct correspondence between each action and spatial grounding, enabling sequentially consistent and accurate execution.}
  \label{fig:action_grounding_comparison}
  \vspace{-3mm}
\end{figure*}

\subsection{Experimental Results}
\noindent\textbf{Overall Performances.}
Table~\ref{tab:main_results} summarizes the performance on \dataset. Among the evaluated models, Gemini-3-Flash leads in end-to-end spatially grounded planning, achieving particularly high Task Success Rates (TSR) in short-horizon tasks when provided with explicit instructions. In comparison, models such as GPT-5.2 and Gemini-2.5 exhibit a significant performance gap compared to Gemini-3-Flash. Furthermore, open-source models like InternVL3.5 and Qwen3-VL struggle even with short-horizon tasks, resulting in substantially lower success rates.

\begin{figure*}[t]

  \centering
  \includegraphics[width=\linewidth]{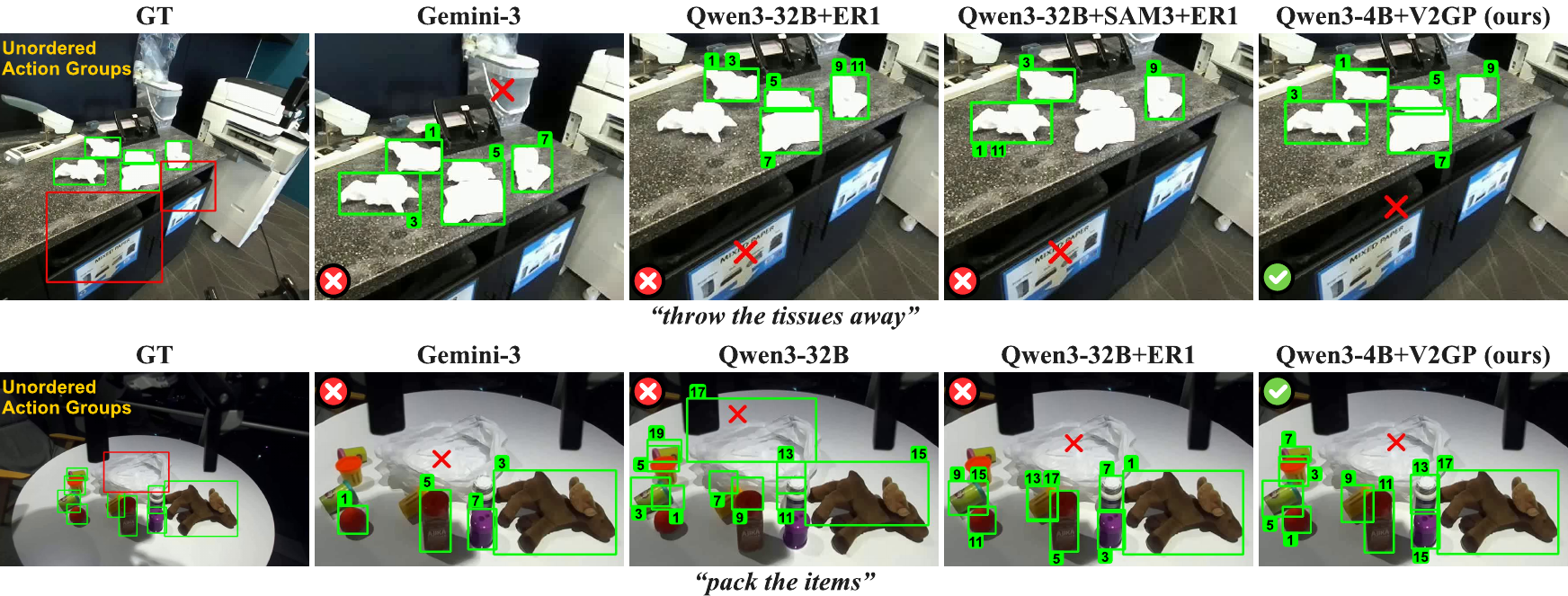}
  \vspace{-4mm}
  \caption{Qualitative comparison of spatially grounded action plans
between ground truth (GT), baseline models, and our method on \dataset.
Green bounding boxes represent grasp predictions,
while red crosses denote placement targets.}
  \label{fig:comparison}
  \vspace{-3mm}
\end{figure*}

\noindent\textbf{Are action plans spatially grounded?} To investigate this, we visualize the spatial grounding of plans generated by decoupled approaches in Fig.~\ref{fig:action_grounding_comparison}. We observe that when using decoupled task planning and spatial grounding (e.g., GPT-5.2+ER-1, Qwen-3+ER-1), the natural language action plans often specify objects ambiguously or redundantly. This misalignment between linguistic description and visual entities prevents the model from correctly grounding all objects required for the task. In the case of SoM (Set-of-Mark) prompting, the task planner frequently fails to select the correct objects, likely due to the high density of visual noise and overlapping prompts in complex real-world settings. In contrast, the task planner fine-tuned with \method consistently achieves accurate spatial localization, successfully mapping each sub-action to its corresponding target.

\noindent\textbf{From Short to Long-Horizon Tasks.} As the task complexity increases from short- to long-horizon, performance across all models degrades significantly. While models may successfully plan short 2--3 steps, maintaining consistent spatial and sequential planning over 9 (long) steps remains a major challenge. Even high-performing models like Gemini-3-Flash see a notable drop in TSR as the horizon lengthens, suggesting that long-horizon task planning still a primary bottlenecks in grounded planning.

\noindent\textbf{From Explicit to Implicit Instructions.} The shift from explicit to implicit instructions further reveals a significant decline in task planning performance for current VLMs. VLMs generally struggle to infer necessary intermediate sub-actions from abstract instructions, an open challenge that demands deeper intent understanding and more advanced reasoning.

\noindent\textbf{Is \method Helpful?} Our results demonstrate that the proposed \method significantly enhances the spatially grounded planning performance of open-source VLMs. By LoRA fine-tuning Qwen3-VL with \method using real-world robot demonstration videos, we observe substantial improvements in Task Success Rate (TSR) across all horizons. For example, in Short-Explicit setting, applying \method improves the TSR of Qwen3-VL-4B from 39.5\% to 58.2\%, even surpassing the performance of its 32B counterpart. Similarly, Qwen3-VL-32B also benefits from \method with particularly strong improvements in challenging settings, achieving 25.9\% TSR in Long-Explicit and 36.3\% in Short-Implicit setting. This suggests that \method contributes to improving complex planning in larger models. We also provide qualitative comparisons in Fig.~\ref{fig:comparison}. Overall, the results indicate that learning from structured sub-action plans with spatial action grounding extracted from demonstration videos allows VLMs to generate plans that are not only sequentially coherent but also grounded in real-world environments.

\subsection{Real-world Experiments}

We further evaluate the executability of generated plans in real-world settings using a Franka Research 3 robot with calibrated Intel RealSense D435i RGB-D cameras in an eye-to-hand configuration. Execution is restricted to predefined action primitives $\{open, grasp, place, close\}$, consistent with our benchmark design. Predicted spatial targets (bounding boxes for grasp and 2D points for placement) are back-projected into 3D space using aligned depth measurements and transformed into the robot base frame via calibrated camera extrinsics. 

\begin{figure*}[t]
  \centering
  \includegraphics[width=0.85\linewidth]{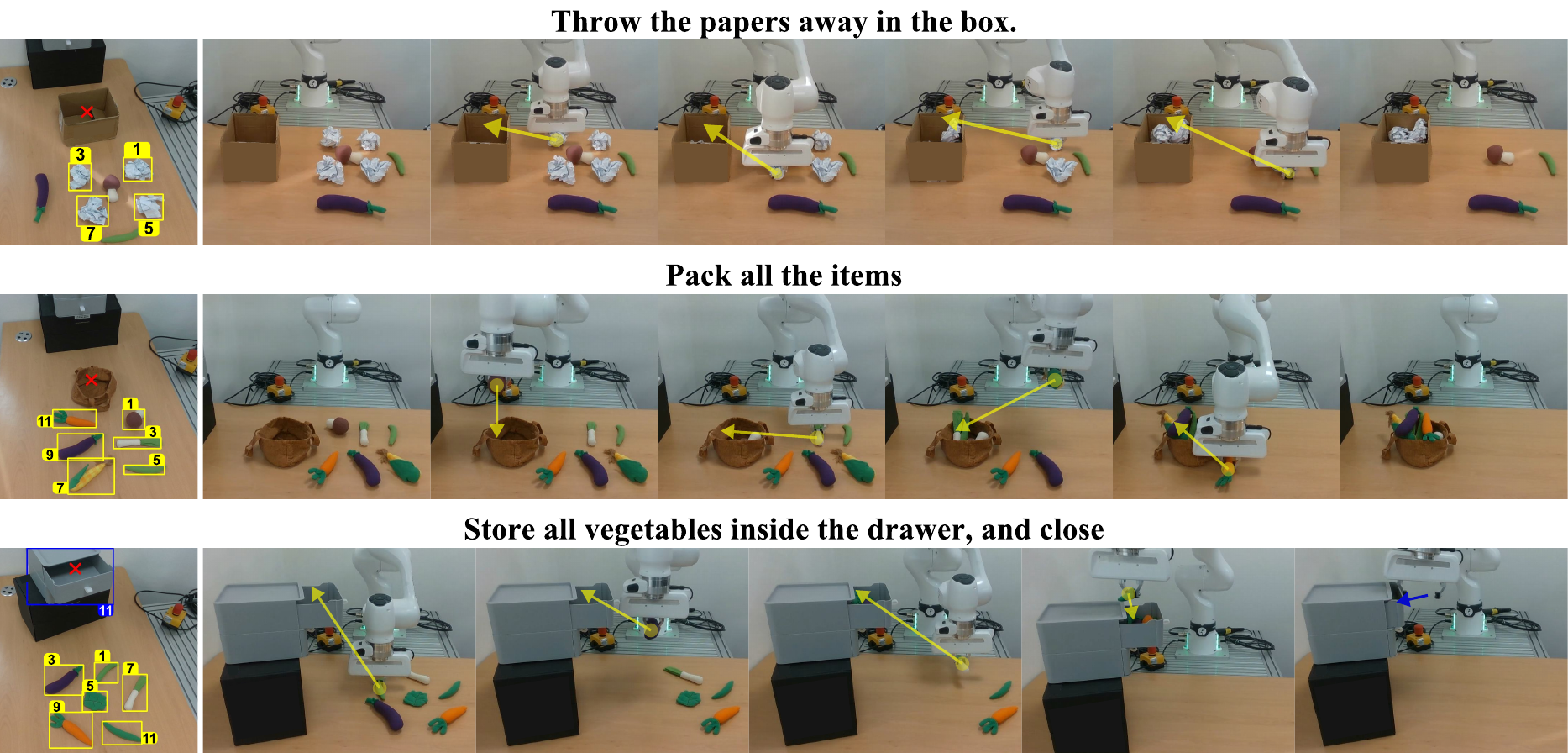}
  \vspace{-1mm}
  \caption{Real-world experimental results from Qwen3-VL-32B+\method. Further results are presented in the video supplement.}
  \vspace{-3mm}
  \label{fig:real_comparison}
\end{figure*}

Specifically, we conduct experiments on 10 tasks in our real-world robotic environment. Table~\ref{tab:qwen32b_comparison} compares the average performance of the Qwen3-VL-32B baseline and Qwen3-VL-32B enhanced with \method. Figure~\ref{fig:real_comparison} presents qualitative examples of executing the generated plans on the real robot, illustrating that the plans produced by \method can be reliably translated into successful robotic executions. Additional demonstrations and corresponding task instructions are provided in the video supplement.

\begin{table}[t]
\caption{Performance comparison on real-world experiments.}
\vspace{-3mm}
\label{tab:qwen32b_comparison}
\begin{center}
\begin{tabular}{c|cc}
\toprule
Model & TSR (\%) & ARR (\%) \\
\midrule
Qwen3-32B & 10.0 & 48.3 \\
Qwen3-32B + \method & \textbf{70.0} & \textbf{93.3} \\
\bottomrule
\end{tabular}
\end{center}
\vspace{-4mm}
\end{table}

\section{Conclusion and Limitations}
This paper addresses the gap between long-horizon task planning (what to do) and spatial grounding (where to act) in existing VLMs. We introduce \taskname and a benchmark, \dataset, that jointly evaluates high-level planning and spatial grounding in real-world environments. Our evaluation of both closed-source and open-source VLMs on \dataset shows that they struggle with spatial grounding in multi-step action planning.
To address this, we propose Video-to-Spatially Grounded Planning (\method), an automated framework that transforms real-world robot demonstrations into spatially grounded planning data. Our experiments show that \method substantially improves spatially grounded planning, enabling more coherent and physically executable action plans.
Despite these advancements, several limitations remain that offer avenues for future research: (1) noisy video signals: sub-action decomposition is sensitive to noisy gripper signals, which can lead to imprecise decomposition. 
(2) hallucination in VLMs: The labeling of interactive objects and spatial grounding targets relies on foundation models, which may occasionally generate hallucinations or misaligned predictions.
(3) scarcity of long-horizon training data: we observe a relative shortage of high-quality real-world robot demonstrations for long-horizon tasks.
Although our framework incorporates a simple filtering/augmentation to mitigate these issues, addressing them remains an important direction for future work.



	{
		\small
		\bibliographystyle{IEEEtran}
		\balance
		\bibliography{IEEEabrv}
	}


%

%
%
%
%
%
%
%

\end{document}

%% file: tables/results3.tex
\begin{table*}[t]
\centering
\small
\setlength{\tabcolsep}{4pt}
\caption{Evaluation results on \dataset\ across varying task horizons (short (1--4), medium (5--8), long (9--26)) and instruction types (explicit and implicit instructions).}
\vspace{-2mm}
\resizebox{\linewidth}{!}{
\begin{tabular}{l|cccccc|cccccc}
\toprule
\multirow{3}{*}{Model}
& \multicolumn{6}{c|}{Task Success Rate (TSR) $\uparrow$}
& \multicolumn{6}{c}{Action Recall Rate (ARR) $\uparrow$} \\

\cmidrule(lr){2-7} \cmidrule(lr){8-13}

& \multicolumn{3}{c}{Explicit}
& \multicolumn{3}{c|}{Implicit}
& \multicolumn{3}{c}{Explicit}
& \multicolumn{3}{c}{Implicit} \\

\cmidrule(lr){2-4} \cmidrule(lr){5-7}
\cmidrule(lr){8-10} \cmidrule(lr){11-13}

& short & medium & long
& short & medium & long
& short & medium & long
& short & medium & long \\

\midrule
\multicolumn{13}{c}{Proprietary VLMs} \\
\midrule
\rowcolor{fftgray}\multicolumn{13}{l}{Decoupled Task Planning + Spatial Grounding} \vspace{0.5mm}\\
GPT 5.2 + ER1
& 41.8 & 13.6 & 8.4
& 31.0 & 10.4 & 3.6
& 54.8 & 48.9 & 46.7
& 45.8 & 42.4 & 38.4 \\

GPT 5.2 + SAM3 + ER1
& 36.2 & 10.1 & 4.9
& 28.0 & 8.7 & 1.4
& 47.6 & 42.4 & 36.9
& 41.9 & 36.2 & 32.1 \\

\rowcolor{fftgray}\multicolumn{13}{l}{End-to-End Spatially Grounded Planning} \vspace{0.5mm} \\

GPT 5.2
& 3.4 & 0.0 & 0.0
& 0.6 & 0.0 & 0.0
& 7.2 & 3.8 & 3.7
& 3.0 & 3.0 & 2.3 \\
Gemini-2.5-Flash
& 20.9 & 11.6 & 5.6
& 10.7 & 6.0 & 5.0
& 28.3 & 28.2 & 29.0
& 21.5 & 20.6 & 22.8 \\
Gemini-3-Flash
& \textbf{67.2} & \textbf{52.0} & \textbf{42.7}
& \textbf{57.1} & \textbf{31.7} & \textbf{17.9}
& \textbf{73.1} & \textbf{71.6} & \textbf{75.1}
& \textbf{67.2} & \textbf{57.8} & \textbf{55.9} \\

\midrule
\multicolumn{13}{c}{Open-Source VLMs} \\
\midrule
\rowcolor{fftgray}\multicolumn{13}{l}{Decoupled Task Planning + Spatial Grounding} \vspace{0.5mm}\\
Qwen3-VL-4B + SoM
& 14.7 & 2.0 & 0.0
& 4.2 & 0.5 & 0.0
& 19.5 & 18.3 & 14.6
& 9.4 & 9.9 & 8.1 \\

Qwen3-VL-4B + ER1
& 37.3 & 9.1 & 2.8
& 18.5 & 4.4 & 0.7
& 49.7 & 39.8 & 32.5
& 33.0 & 26.4 & 21.0 \\

Qwen3-VL-4B + SAM3 + ER1
& 35.6 & 6.1 & 2.8
& 18.5 & 2.2 & 0.7
& 48.6 & 33.4 & 30.6
& 31.8 & 23.7 & 20.2 \\

Qwen3-VL-32B + ER1
& 40.1 & 12.6 & 8.4
& 29.8 & 7.1 & 2.9
& 52.0 & 47.0 & 45.4
& 45.5 & 36.3 & 35.6 \\

Qwen3-VL-32B + SAM3 + ER1
& 36.7 & 11.6 & 4.9
& 29.2 & 6.0 & 0.7
& 48.8 & 41.6 & 42.4
& 43.7 & 31.6 & 33.7 \\
\rowcolor{fftgray}\multicolumn{13}{l}{End-to-End Spatially Grounded Planning} \vspace{0.5mm}\\
InternVL3.5-8B
& 0.0 & 0.0 & 0.0
& 0.6 & 0.0 & 0.0
& 0.3 & 0.3 & 0.6
& 0.6 & 0.3 & 0.0 \\


\midrule

Qwen3-VL-4B
& 39.5 & 18.2 & 5.6
& 22.6 & 6.0 & 1.4
& 50.9 & 43.3 & 36.7
& 35.0 & 27.4 & 23.0 \\

Qwen3-VL-4B + \method

& \textbf{58.2} & \textbf{33.8} & \textbf{21.7}
& \textbf{31.0} & \textbf{13.1} & \textbf{7.9}
& \textbf{66.8} & \textbf{60.0} & \textbf{53.5}
& \textbf{42.9} & \textbf{35.3} & \textbf{31.9} \\

\midrule

Qwen3-VL-32B
& 42.9 & 30.8 & 14.0
& 23.8 & 12.0 & 2.9
& 55.6 & 55.7 & 54.2
& 39.9 & 37.7 & 33.4 \\

Qwen3-VL-32B + \method
& \textbf{58.8} & \textbf{33.8} & \textbf{25.9}
& \textbf{36.3} & \textbf{14.8} & \textbf{12.9}
& \textbf{69.1} & \textbf{60.0} & \textbf{60.3}
& \textbf{49.0} & \textbf{39.0} & \textbf{38.0} \\

\bottomrule
\end{tabular}
}
\vspace{-1mm}
\label{tab:main_results}
\end{table*}